\documentclass{article}

\usepackage{microtype}
\usepackage{graphicx}
\usepackage{subfigure}
\usepackage{booktabs} % for professional tables
\usepackage[hyphens]{url}

\usepackage{hyperref}

\usepackage[accepted]{icml2024}

\usepackage{amsmath}
\usepackage{amssymb}
\usepackage{mathtools}
\usepackage{amsthm}

\usepackage[capitalize,noabbrev]{cleveref}

\theoremstyle{plain}

\theoremstyle{definition}

\theoremstyle{remark}

\usepackage[textsize=tiny]{todonotes}

\icmltitlerunning{Social Choice Should Guide AI Alignment}

\begin{document}

\twocolumn[
\icmltitle{Social Choice Should Guide AI Alignment \\ in Dealing with Diverse Human Feedback}



\icmlsetsymbol{equal}{*}

\begin{icmlauthorlist}
\icmlauthor{Vincent Conitzer}{cmu,oxf}
\icmlauthor{Rachel Freedman}{chai}
\icmlauthor{Jobst Heitzig}{pots}
\icmlauthor{Wesley H. Holliday}{philucb}
\icmlauthor{Bob M. Jacobs}{ghent}
\icmlauthor{Nathan Lambert}{allen}
\icmlauthor{\quad Milan Mossé}{philucb}
\icmlauthor{\quad Eric Pacuit}{umd}
\icmlauthor{\quad Stuart Russell}{chai}
\icmlauthor{\quad Hailey Schoelkopf}{elai}
\icmlauthor{Emanuel Tewolde}{cmu}
\icmlauthor{William S. Zwicker}{uni,msc}
\end{icmlauthorlist}

\icmlaffiliation{cmu}{Foundations of Cooperative AI Lab, Computer Science Department, Carnegie Mellon University, Pittsburgh, USA}
\icmlaffiliation{oxf}{Institute for Ethics in AI, University of Oxford, Oxford, United Kingdom}
\icmlaffiliation{chai}{Center for Human-Compatible AI, Department of Electrical Engineering and Computer Sciences, University of California, Berkeley, USA}
\icmlaffiliation{pots}{Potsdam Institute for Climate Impact Research, Potsdam, Brandenburg, Germany}
\icmlaffiliation{philucb}{Department of Philosophy, University of California, Berkeley, USA}
\icmlaffiliation{ghent}{Department of Philosophy and Moral Sciences, Ghent University, Ghent, Begium}
\icmlaffiliation{allen}{Allen Institute for AI, Berkeley, California, USA}
\icmlaffiliation{umd}{Department of Philosophy, University of Maryland, College Park, USA}
\icmlaffiliation{elai}{EleutherAI}
\icmlaffiliation{uni}{Department of Mathematics, Union College, Schenectady, USA}
\icmlaffiliation{msc}{Murat Sertel Center for Advanced Economic Studies, Istanbul Bilgi University, Istanbul, Turkey}

\icmlcorrespondingauthor{Vincent Conitzer}{conitzer@cs.cmu.edu}

\icmlkeywords{Machine Learning, ICML}

\vskip 0.3in
]



\printAffiliationsAndNotice{}  % leave blank if no need to mention equal contribution

\begin{abstract}
Foundation models such as GPT-4 are fine-tuned to avoid unsafe or otherwise problematic behavior, such as helping to commit crimes or producing racist text. One approach to fine-tuning, called \textit{reinforcement learning from human feedback}, learns from humans' expressed preferences over multiple outputs. Another approach is \textit{constitutional AI}, in which the input from humans is a list of high-level principles. But 
how do we deal with potentially diverging input from humans? How can we aggregate the input into consistent data about ``collective'' preferences or otherwise use it to make collective choices about model behavior? 
In this paper, we argue that the field of {\em social choice} is well positioned to address these questions, and we discuss ways forward for this agenda, drawing on discussions in a recent workshop on 
Social Choice for AI Ethics and Safety held in Berkeley, CA, USA in December~2023.
\end{abstract}

\section{Introduction}
\label{sec:intro}

Recently, \textit{reinforcement learning from human feedback} (RLHF) has become the primary strategy that leading AI companies such as OpenAI~\cite{openai2023gpt4}, Anthropic~\cite{anthropic2023}, Meta~\cite{meta2023llama2}, and Google~\cite{google2023} use to make pretrained large language models (LLMs) more capable and controllable~\cite{christiano2017deep, ziegler2019fine} and to align them with human values. However, RLHF faces many limitations and challenges~\citep{casper2023open, lambert2023alignment}, including unrepresentative data~\cite{prabhakaran2021releasing,feffer2023moral}, unrealistic models of human decision-making~\cite{hong2022sensitivity, freedman2021choice, siththaranjan2023distributional, lambert2023history}, and insufficient modeling of human diversity~\cite{kirk2023personalisation,freedman2023active} which may lead to political bias \cite{MotokiNR23,Rozado24}. 
We propose that ideas from social choice theory~\citep{arrow2012social,Fishburn1973,Kelly1988,Brandt15:Handbook}---e.g., concerning whose preferences should be integrated into decisions and how this should be done---are needed to solve many of these open problems.

While models that are solely pretrained on internet data may produce repetitive or harmful text, RLHF trains models to follow instructions~\cite{ouyang2022training} and produce helpful and ``harmless'' outputs~\cite{bai2022training} based on human judgments. 
RLHF gathers example outputs from a pretrained LLM or examples written by humans. 
Next, humans are asked to select the outputs that best meet specified criteria (such as being ``helpful'' or ``unbiased''). 
These judgments, often called \textit{preferences}, are then used to fine-tune the LLM. 
From a social choice perspective, this method raises several critical questions: Which humans are asked to judge outputs? What criteria do they use? How are their judgments combined? And how do their expressed judgments relate to their actual preferences?

\textit{Constitutional AI} (CAI), which involves reinforcement learning from AI feedback (RLAIF), is an alternative addressing some of these questions~\cite{Bai22:Constitutional}. 
Humans produce a ``constitution'' that specifies principles that the LLM is trained to align with. 
However, one must still decide who has input on the constitution and how it is constructed. 
\citet{Bai22:Constitutional} construct it ``in a fairly ad-hoc way [\dots] for research purposes'', but developing safe and ethical AI requires a more principled approach as in  \citet{ganguli2023} or \citet{openai2024democratic}. How then should one aggregate diverse preferences into a representative constitution?

Social choice theory has long studied similar questions.
\textbf{This position paper argues that methods from social choice should be applied to address questions such as which humans should provide input, what type of feedback should be collected, and how it should be aggregated and used.}  
By taking into account the lessons from social choice theory, one can avoid na\"ive mistakes and reinventing the wheel, while leveraging feedback to address challenging design problems~\citep{dobbe2021hard}.
We also highlight areas in which new work is required to extend social choice to new problems unique to training safe and ethical AI.

There are several advantages to addressing the above problems in a principled way. First, it will likely result in a fairer system,  taking into account the input of a broader group of people. Second, it promises to give generally more accurate feedback about questions of truthfulness, as suggested by a significant body of literature on ``epistemic democracy''---voting to settle questions about facts \cite{pivato2017epistemic}. Intuitively, having input from diverse people makes it less likely that something important is missed. Third, it will likely result in broader buy-in into the system. 

One may have concerns about this approach; for example, will feedback from diverse people be inconsistent and result in inconsistent system behavior? Social choice theory provides examples where na\"ive aggregation of preferences or judgments leads to seemingly irrational collective choices, such as cyclical preferences \cite{Schwartz2018} or logically inconsistent conclusions \cite{ListPettit2002}. Then again, social choice theory also provides the tools for thinking about such issues and preventing them.

Social choice is not new to computer scientists; {\em computational social choice}~\cite{Brandt15:Handbook} is a well-studied topic, with a biennial workshop since 2006. However, while many of those researchers affiliate with the AI community, there has not yet been much work connecting computational social choice to the alignment of modern AI systems. 

In the following, we first give background on value alignment, RLHF, and social choice.  Then we discuss several questions at their intersection.  We think that significant further research is needed to answer these questions well and that good answers are needed in order to  build AI systems in a responsible way based on potentially diverging feedback from multiple stakeholders. In contrast, ad-hoc approaches may result in systems that fail to represent their stakeholders well, that marginalize significant groups of stakeholders, and that create a basis for conflict between groups of people or the multiple AI systems that represent them.

\section{Background}
\label{sec:background}

Our proposed research agenda requires background from so far mostly disjoint communities.  Readers familiar with some of the following can skip the corresponding subsections.

\subsection{Value Alignment}
\label{sec:value alignment}

As AI systems become more capable, it becomes critical that they act in alignment with human and societal values~\cite{gabriel2020artificial}. Many approaches to \textit{value alignment} exist, such as, e.g., formal games in which AI agents must align with humans to solve them~\cite{shah2020benefits}, empirics on the relation between neural network activations and morally relevant output features~\cite{zou2023representation}, and evaluations of the ethical behavior of LLMs~\cite{pan2023rewards}. RLHF is a particularly popular but so far limited approach.

\subsection{Reinforcement Learning from Human Feedback}
\label{sec:background rlhf}

RLHF begins with generating and evaluating a dataset of model outputs $\mathcal{Y}$. 
In vanilla RLHF, humans are then shown paired completions $\{y_0,y_1\}\in\mathcal{Y}\times\mathcal{Y}$ to prompts $x\in\mathcal{X}$ 
and asked which output $y\in\{y_0,y_1\}$ they prefer~\citep{christiano2017deep,lee2021pebble}. 
Other RLHF variants ask humans to rank or provide scores for groups of outputs~\cite{ziegler2019fine,ouyang2022training}, and many additional variations exist~\citep{wu2023fine}.

The next step is to fit a parameterized reward model $\varrho_\theta\colon \mathcal{Y}\rightarrow\mathbb{R}$. 
For LLMs, $\varrho_\theta$ is typically a neural network with weights $\theta$. 
RLHF methods assume that there is a ground-truth reward function $\varrho_{\theta^*}$ that the human preferences reflect (up to noise). 
The reward model 
is then optimized to match the likelihoods of the human preferences observed in the data. If the training data comes from diverse sources, this implicitly amounts to a rather intransparent or flawed form of preference aggregation \citep{siththaranjan2023distributional,Xu23iia,Chakraborty24,ge2024axioms}.

The final step is to use reinforcement learning to train a policy that maximizes rewards from the reward model. 
This involves many design decisions—which RL algorithm to use, how to regularize the updates, and whether to gather further online feedback during training. 
See \citet{uc2023survey} for a survey on the use of RL to train LLMs.

\subsection{Alternate Preference-Based Fine-Tuning Objectives}
\label{sec:alt_po}
Since common RL methods like PPO~\citep{schulman2017proximal} can be unstable, novel techniques for optimizing a model based on collected preference data have been proposed. \citet{rafailov2023direct} introduce Direct Preference Optimization (DPO), which recasts RLHF to converge to the best solution by directly optimizing a loss on the preference label dataset, rather than sampling online from the LLM policy or training an explicit reward model. 
Another variant emerged to remove the dependency on pairwise data. 
\citet{ethayarajh2023halos} propose a loss function termed Kahneman--Tversky Optimization (KTO) that enables learning a policy from \textit{unpaired} preferences. 
The authors further claim that the effectiveness of various losses for RLHF depends on the properties they share with proposed human utility functions \citep{prospecttheory}.

\subsection{Constitutional AI}
\label{sec:background const AI}

\citet{Bai22:Constitutional} further explore the design space with {\em Constitutional AI (CAI),} which relies on {\em RL from AI Feedback (RLAIF).} 
RLAIF is a larger set of techniques for using AI to augment or generate feedback data, including pairwise preferences~\citep{lee2023rlaif, sharma2024critical, castricato2024suppressing}.  
By employing a human-written set of principles, which they term a \textit{constitution}, \citet{Bai22:Constitutional} use a separate LLM to generate artificial preference and instruction data used for fine-tuning.
A constitution $\mathcal{C}$ is a set of written principles indicating specific aspects to focus on during a critique phase.
The instruction data is curated by repeatedly sampling a principle $c_i \in \mathcal{C}$ and asking the model to revise its latest output $y^i$ to the prompt $x$ to align with $c_i$. 
This yields a series of instruction variants $\{y^0, y^1, \cdots, y^n\}$ from the principles  $\{c_{0}, c_{1}, \cdots, c_{n-1}\}$ used for critique.
The final data point is the prompt $x$ together with the final completion $y^n$, for some $n$. 

The preference data is constructed in a similar, yet simpler way by using a subset of principles from $\mathcal{C}$ as context for a feedback model.
The feedback model is presented with a prompt $x$, a set of principles $\{c_0, \cdots, c_n\}$, and two completions $y_0$ and $y_1$ labeled as answers (A) and (B) from a previous RLHF dataset.
The feedback models' probability of outputting either (A) or (B) is recorded as a training sample for the reward model, as discussed in \Cref{sec:background rlhf}.

\subsection{Social Choice}
\label{sec:background sc}

Modern social choice theory began in the 1950s with Arrow's Impossibility Theorem \cite{Arrow1951,McLean1995}. 
Arrow considered the problem of aggregating multiple individuals' preferences---in his case rankings---into a social preference, subject to some desiderata. In particular, he required the aggregation function to be defined for any family of individual preferences; that the social preference relation be complete and transitive (it is then called a \textit{social welfare function}); that the social preference between alternatives $A$ and $B$ should depend only on individual preferences between $A$ and $B$ (Independence of Irrelevant Alternatives, IIA); and that unanimous individual preference for $A$ over $B$ should imply social preference for $A$ over $B$. Arrow proved that the only aggregation functions satisfying these desiderata are \textit{dictatorships}: there is one individual $d$ such that no matter what others prefer, if $d$ prefers $A$ to $B$, then the social preference ranks $A$ above $B$ as well.\footnote{\citet{Mishra2023} applies Arrow's Theorem to RLHF.} A similar theorem (see \citealt{Taylor2005}, \S~1.3) holds for {\em social choice functions} where, instead of asking for a social ranking, we ask for just a single winner, or even just a set of choice-worthy alternatives.

Arrow's Theorem stimulated a huge literature exploring the consequences of weakening his desiderata (see, e.g., \citealt{Campbell2002}, \citealt{HP2020}). The general takeaway is that for ordinal preference aggregation, in order to avoid dictatorships, oligarchies and vetoers, one must weaken IIA and allow the social preference between $A$ and $B$ to depend in part on preferences involving other alternatives. This allows for many alternative methods of aggregating individual preferences (see, e.g., \citealt{Brams2002,Zwicker2016,Pacuit2019} and the voting methods in the \href{https://pref-voting.readthedocs.io/en/latest/}{Preferential Voting Tools} library). Figure \ref{Fig:diffagg} gives an example in which three well-known methods disagree. The costs and benefits of these and other methods are systematically studied from various angles (axiomatic, computational, empirical, etc.) in social choice theory.

\begin{figure}
\begin{center}\begin{tabular}{cccccc}
$4$&$4$ & $9$  & $4$ & $2$\\\hline 
$A$&$A$ & $B$  & $C$ & $C$\\ 
$B$&$C$ & $C$  & $A$ & $B$\\ 
$C$&$B$ & $A$ & $B$ & $A$
\end{tabular} \begin{tabular}{l}
Borda Count: $CBA$ \\
Instant Runoff: $ABC$ \\
Ranked Pairs: $BCA$
\end{tabular}
\end{center}
\caption{Individual rankings on the left (4 voters say $ABC$, 4 say $ACB$, etc.)~lead to different aggregations on the right, depending on the aggregation rule. Borda Count gives an alternative 0, 1, or 2 points for each voter who ranks it last, second, or first, respectively; alternatives are then ordered by score. Instant Runoff ranks $C$ last since $C$ has the fewest first-place rankings; after removing $C$, $B$ has the fewest first-place rankings, so $B$ is in second and $A$ first. For Ranked Pairs, notice there is a \textit{majority cycle}: a majority of voters prefer $A$ to $B$, a majority prefer $B$ to $C$, and a majority prefer $C$ to $A$; the smallest margin of victory is for $A$ over $B$, so we drop this majority preference, yielding $BCA$.}\label{Fig:diffagg}
\end{figure}

Since Arrow, social choice theory has grown to study aggregation not only of individuals' preferences, both ordinal and cardinal \cite{Daspremont2002}, but also of their \textit{approvals} of alternatives \cite{LaslierSanver2010}, \textit{grades} given to alternatives \cite{Balinski2010}, \textit{judgments} about propositions \cite{GrossiPigozzi2022}, \textit{subjective probabilities} for propositions \cite{DietrichList2016}, and other types of objects \cite{Rubinstein1986}. Recent work has also started to explore social choice methods for RLHF \cite{Song24PRO,Swamy24Minimaximalist,DaiF24,ge2024axioms}. In the following, we discuss some of the aggregation problems that might arise in the context of AI alignment.

\section{What Are the Collective Decision Problems and their Alternatives in this Context?}
\label{sec:alternatives}

If we want to use social choice methods for aligning AI systems, we first need to specify what the concrete options/objects are, then  collect preferences over them, and finally make actual or simulated collective choices between them. These options are called {\em alternatives.}
In some contexts, the set of alternatives is easy to comprehend and enumerate, e.g. candidates for a position. In other settings, there are exponentially many alternatives, but the set is still easy to comprehend, e.g., if each of $n$ propositions must be either accepted or rejected~\cite{Lang07:Vote}.

For AI alignment, it is harder to see how best to determine the set of alternatives for evaluation. It could be the set of all possible {\em parameterizations} of a given network architecture, but this would surely be conceptually intractable.

For an LLM, most RLHF approaches ask the evaluator to choose between a small, explicit set of alternative {\em responses} to some prompt, handwritten or sampled from the pretrained LLM. Alternately, we could consider {\em all} possible responses as alternatives. Evaluators could then indicate their preference by providing the preferred response themselves. Such exemplars are often used for fine-tuning and can be used to learn evaluators' preferences and generate responses that well-represent them \cite{fish2023generative}. 
While this does not address questions about how to generalize beyond a single prompt, it is a useful way of conceptualizing the alternatives. 

One might conceive of the alternatives as {\em probability distributions} over responses; LLMs are anyway configured to respond stochastically. This might be desirable not only for creativity but also to promote fairness and representativeness of responses. In response to a controversial question, fairness might militate against always giving the same answer, as any one answer will inevitably omit some relevant considerations on one side of a debate. There is a large literature on social choice rules that output probability distributions. Their input could be the evaluators' stated preferences between distributions (\citealt{Fishburn1973}, Ch.~18), or stated preferences between plain alternatives \cite{Brandt2017}, since evaluators might have difficulties comparing probability distributions. Indeed, the type of objects chosen by a social choice rule (e.g., distributions over responses) need not match the type of objects about which individuals state preferences (e.g., responses). 

Multi-winner rules \cite{faliszewski2017multiwinner,elkind2017properties} form a middle ground between deterministic single-winner rules and probabilistic rules. They pick a small, predetermined number of answers that best reflect what the voters want, which could be combined into a single response that lists these answers as bullet points to provide the user with a representative overview of possible answers. 

\section{Who Provides the Human Feedback?}

Let us assume a \textit{stakeholder population} of people who will be affected by an AI system and whose preferences would therefore ideally be considered in aligning the AI.\footnote{Some stakeholders, such as small children and non-human animals, whose preferences we cannot easily elicit might need to be represented by dedicated others.} When it is infeasible to elicit feedback from all stakeholders, we must select a smaller group to query. One can try to select a suitably representative subset such that the alignment obtained using their feedback sufficiently approximates the alignment that would be obtained from all stakeholders' feedback. 
Here one could draw on ongoing work in social choice theory on how to select citizens' assemblies that are representative of a full population (e.g.,~\citealt{Flanigan21:Fair,Landemore22:Citizens}), as well as work in statistics on efficient stratified sampling (e.g., \citealt{Meng2013}). 

Alternatively, one could let stakeholders vote on their representatives, e.g., with a voting procedure designed to elect proportionally representative assemblies (see, e.g., Ch.~4 of \citealt{LacknerSkowron2023}). Stakeholders might also delegate their feedback rights to others (who may in turn delegate, etc.), as in \textit{liquid democracy} (see \citealt{Paulin2020}).

Work up to now  has used evaluator recruitment methods such as Mechanical Turk \citep{Freedman20:Adapting, bai2022training}; Upwork, Scale AI, or Lionbridge \citep{stiennon2020learning, ziegler2019fine}; and purpose-built platforms \citep{noothigattu2018voting}. We believe this component of the RLHF pipeline deserves a more in-depth discussion, including one informed by social choice theory.

\section{What Is the Format of Human Feedback?}
\label{sec:HF format}

As discussed, human feedback for AI systems can have various forms; which of these are most natural and useful?
Here, we can draw on a significant literature on
{\em
preference elicitation} (see, e.g.,~\citealt{Sandholm06:Preference}), studying how best to query agents for their preferences in several domains.\footnote{Incidentally, \citet{li2023eliciting} propose to use LLMs for this.}
This literature is closely tied to that of {\em communication complexity} (e.g.,~\citealt{Kushilevitz97:Communication}), which is concerned with minimizing the number of bits needed to communicate something. Both of these topics have also been studied in voting settings~\cite{Conitzer02:Elicitation,Conitzer05:Communication,Service12:Communication}.

\subsection{Multiple Format Options}
\label{sec:multiple formats format}

In general, we want the type of input or feedback that we ask of humans to be 
(1) natural to give, 
(2) informative about their preferences and values, 
and
(3) of a type that can be used to align AI systems. 
For example, having humans comment on an AI output in an open-ended text box may satisfy 1 and 2, but not 3 (at least, with current methods). Having them sort responses alphabetically may satisfy 1 and 3, but not 2. Having them directly rank neural networks based on inspecting their weights may satisfy 3 but not 1 or 2.
Needless to say, different choices for the type of input may lead to differently aligned systems and have different behavioral effects on humans (cf.\ \Cref{sec:behav aspects}). 

Perhaps one should  let individuals {\em choose} the format in which they give input or feedback. In traditional social choice this is uncommon---though there may be some flexibility in how preferences are expressed (e.g., allowing voters to only rank a few alternatives rather than all~\cite{HalpernKPTW23}, or to give numerical ratings instead of rankings) and some variety in the interaction mechanism (e.g., one can vote for candidates individually, or pull a lever to vote for all candidates of a single party at once). Going forward, however, it is easy to imagine giving evaluators choices from a range of different ways to give their input. For example, the input could be individual responses, whole dialogue sessions, longterm interactions with the same user, or published guiding principles. The feedback could be expressed as approval/disapproval votes, pairwise comparisons of alternatives, full or partial rankings of the alternatives, giving precise or imprecise ratings of the form ``I rate A between 7 and 9'', or even in free-form verbal feedback that an LLM can interpret as some formal preference data such as a partial ordering. Moreover, the evaluation could be on various aspects of the system's behavioral patterns, as done in fine-grained RLHF~\citep{wu2023fine}, or in RLHF that optimizes for multiple attributes~\citep{dong2023steerlm} such as helpfulness, humor, toxicity, etc. 
\Cref{sec:dealwithdivfb} discusses how we may process all this heterogeneous data.

\subsection{Processing Diverse and Informal Feedback}
\label{sec:dealwithdivfb}

Recall that in RLHF, human feedback is typically used to train a reward (or ``preference'') model, which maps any possible AI system response to a numerical rating. 
The concept of reward models could also be used to convert an evaluator's diverse input into a common form, in order to then aggregate it 
with other evaluators' input. 

First, an {\em individual evaluation interpretation model} $\phi$ could be trained to map a tuple of inputs of the form $(x,\mathcal{Y},f_i,e,y)$ to a numerical evaluation $r$. 
As before, $x$ represents a prompt to the AI system, $\mathcal{Y}$ the set of possible AI responses, and $y \in \mathcal{Y}$ a particular response. 
Vector $f_i$ represents the relevant features of a certain evaluator $i$, and $e$ is a language representation of $i$'s feedback on possible responses $\mathcal{Y}$ to $x$, 
containing preference- and evaluation-related statements of various types (see \Cref{sec:multiple formats format}). 
In practice, $\phi$ would likely be based on an LLM pretrained to understand the texts $x$, $\mathcal{Y}$, $e$, and $y$, that is then fine-tuned to the interpretation task described above. Then the output $r = \phi(x,\mathcal{Y},f_i,e,y)$ of $\phi$ is a numerical rating of $y$ given by evaluator $i$ that is trained to be 
(approximately) consistent with the verbal evaluation $e$ of that evaluator. We note that this task can be seen as a form of meta-learning.

One could then use the trained evaluation interpretation model $\phi$ to train another model---an {\em individual preference model} $\psi$---that skips verbal evaluations and directly maps inputs $(x,\mathcal{Y},f_i,y)$ to ratings $r = \psi(x,\mathcal{Y},f_i,y)$. Namely, any tuple $(x,\mathcal{Y},f_i,e)$ can be converted into supervised training data $\big( (x,\mathcal{Y},f_i,y), \phi(x,\mathcal{Y},f_i,e,y) \big)_{y \in \mathcal{Y}}$ for $\psi$, containing simulated ratings $r=\phi(x,\mathcal{Y},f_i,e,y)$. 
The hope is that the individual preference model $\psi$ would be able to simulate the rating of any evaluator (represented by their features $f_i$), 
as long as the evaluator, prompt, and response set come from the same distribution as the one $\psi$ was trained on. 
Similar to the preference models used in current RLHF, $\psi$ could finally be used to fine-tune the actual AI system or steer its behavior in real time. 
In fact, if the evaluators' features $f_i$ are omitted in the training process sketched above, $\psi$ reduces to the standard preference model as is already used in RLHF. This is vulnerable to, for example, evaluators that strategically misreport \cite{siththaranjan2023distributional}, or to issues that arise from a disproportionate representation in the set of evaluators. 
We add evaluators' features $f_i$ to $\psi$ so that preferences can later be aggregated in a transparent and deliberate manner via an additional {\em social choice step}, as we discuss next. 

\section{How Do We Incorporate Diverse Individual Feedback?}
\label{sec:incorporating div fb}

Here we sketch several variants of two approaches for including diverse input or feedback into AI systems in a consistent way using methods from social choice theory. 
The first proposes adding an additional {\em preference aggregation step} somewhere during training, thereby turning RLHF into RLCHF: Reinforcement Learning from Collective Human Feedback. 
The second approach instead proposes adding an additional {\em simulated collective decision step} somewhere in the training or the system's real-time decision procedure, similar to \citet{bakker2022fine} and \citet{jarrett2023language}. 

\begin{figure*}[t]
\centering
\includegraphics[width=0.7\textwidth]{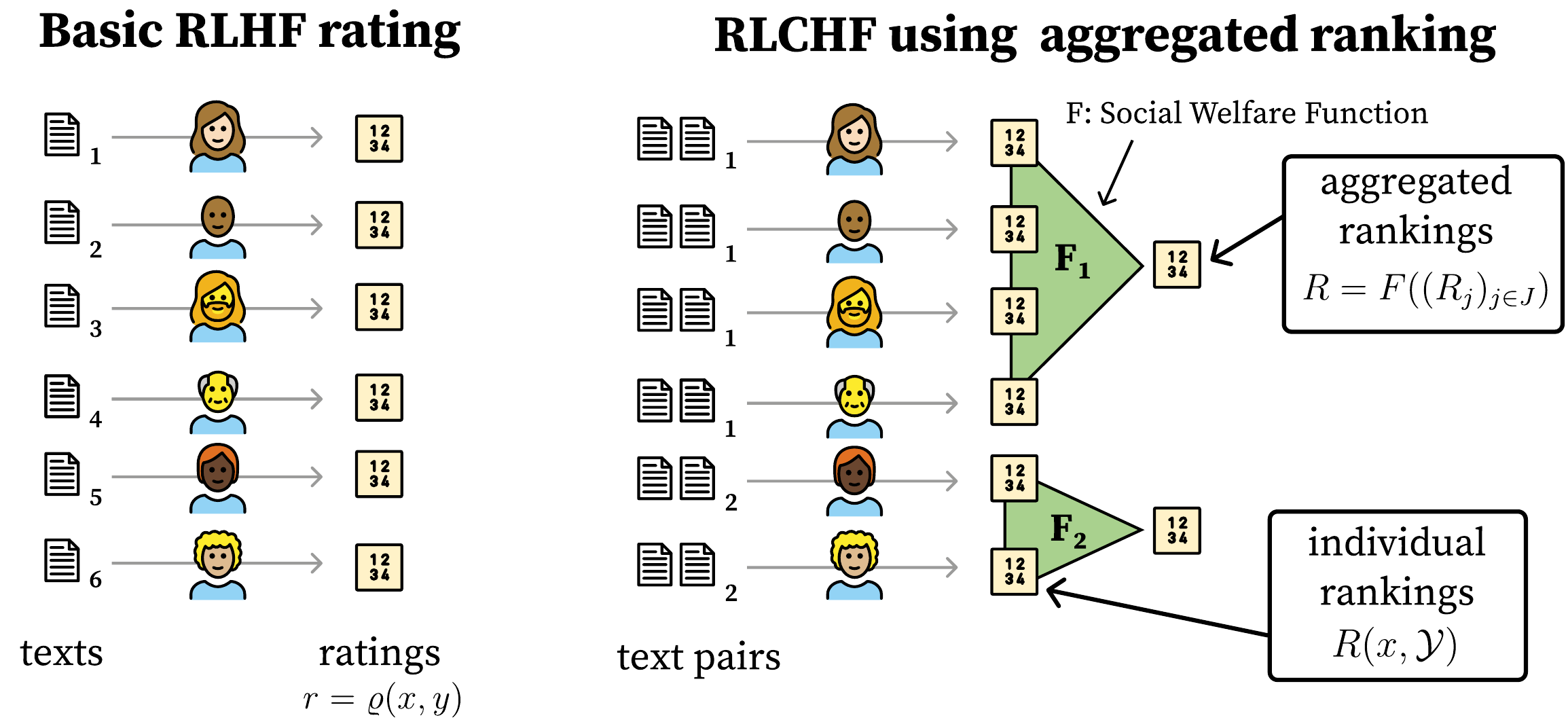}
\caption{\textbf{RLCHF using aggregated rankings}. The core addition to the standard RLHF process is the call-out of an explicit social welfare function, $F$, which determines how preferences are aggregated.}
\label{fig:rlchf}
\end{figure*}

\subsection{Proposal: Reinforcement Learning from Collective Human Feedback (RLCHF)}
\label{sec:rlchf}

Preference aggregation could be incorporated into RLHF in several ways, early on or rather late.
For simplicity, assume a base version of \emph{rankings-based} RLHF that (1) takes a database 
of prompts $x$ together with corresponding sets of possible responses $\mathcal{Y}$, (2) asks one associated evaluator $i(x,\mathcal{Y})$ to provide a ranking $R(x,\mathcal{Y})$ of the elements of $\mathcal{Y}$, (3) turns this ranking into $|\mathcal{Y}|$ many data points for training a common preference model $\varrho$ that produces numerical ratings $r=\varrho(x,y)$, and (4) uses these ratings as rewards in fine-tuning the actual 
LLM via reinforcement learning.

The earliest point one may introduce preference aggregation is between steps (2) and (3). Instead of a single evaluator, we may ask the members of a jury $J(x,\mathcal{Y})$ to provide individual rankings $R_j$. Using some ordinal social welfare function~$F$, those rankings can then be aggregated into a collective ranking $R = F((R_j)_{j\in J})$ to be used in step (3). We call this ``RLCHF using aggregated rankings'' (Fig.~\ref{fig:rlchf}).  

\begin{figure*}[t]
\centering
\includegraphics[width=0.8\textwidth]{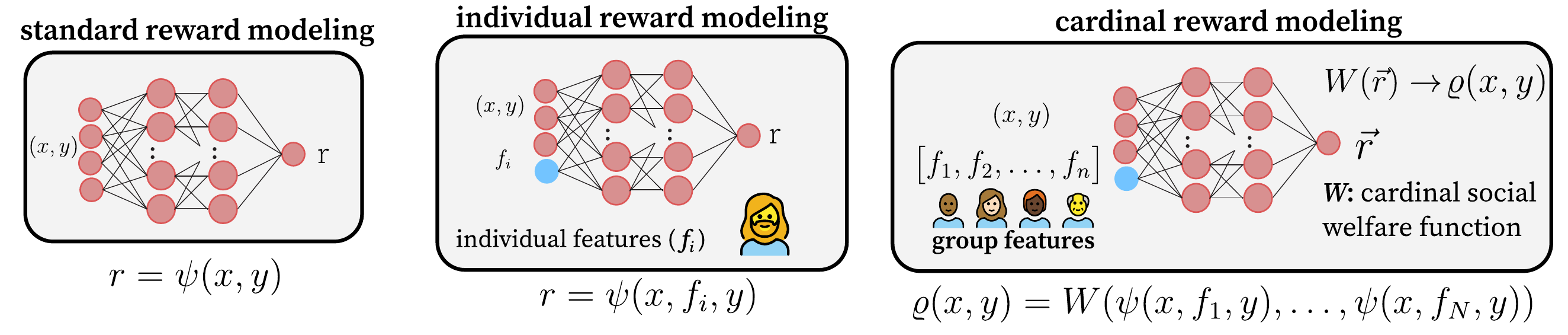}
\caption{\textbf{RLCHF using evaluator features and aggregated ranks}. We show how an individuals' features can be used as an additional input to reward models within the RLHF process.}
\label{fig:rlchf-f}
\end{figure*}

Alternatively, one could use {\em cardinal} preference aggregation between steps (3) and (4). For this, change step (3) so that a model of {\em individual} preferences is trained, mapping pair $(x,\mathcal{Y})$ and evaluator $i$ with features $f_i$ to predicted ratings $r_i = \psi(x,f_i,y)$. Also, generate a large sample of feature vectors $f_1,\dots,f_N$ that is representative of the stakeholder population. Then a {\em cardinal} social welfare function $W$ can be used to aggregate simulated individual ratings into a social rating $\varrho(x,y) = W(\psi(x,f_1,y),\dots,\psi(x,f_N,y))$ which can be used in step (4). We call this ``RLCHF using evaluator features and aggregated ratings'' (Fig.~\ref{fig:rlchf-f}).

\subsection{Proposal: Simulated Collective Decisions}
\label{sec:sim coll decs}

RLCHF, as described above, keeps the reinforcement learning step that requires numerical rewards, and it uses ordinal or cardinal preference aggregation to produce those rewards for all possible responses $y\in \mathcal{Y}$. A different approach would replace reinforcement learning by something else and introduce social choice methods in the form of simulated collective decisions rather than preference aggregation.

\begin{figure*}[t]
\centering
\includegraphics[width=0.9\textwidth]{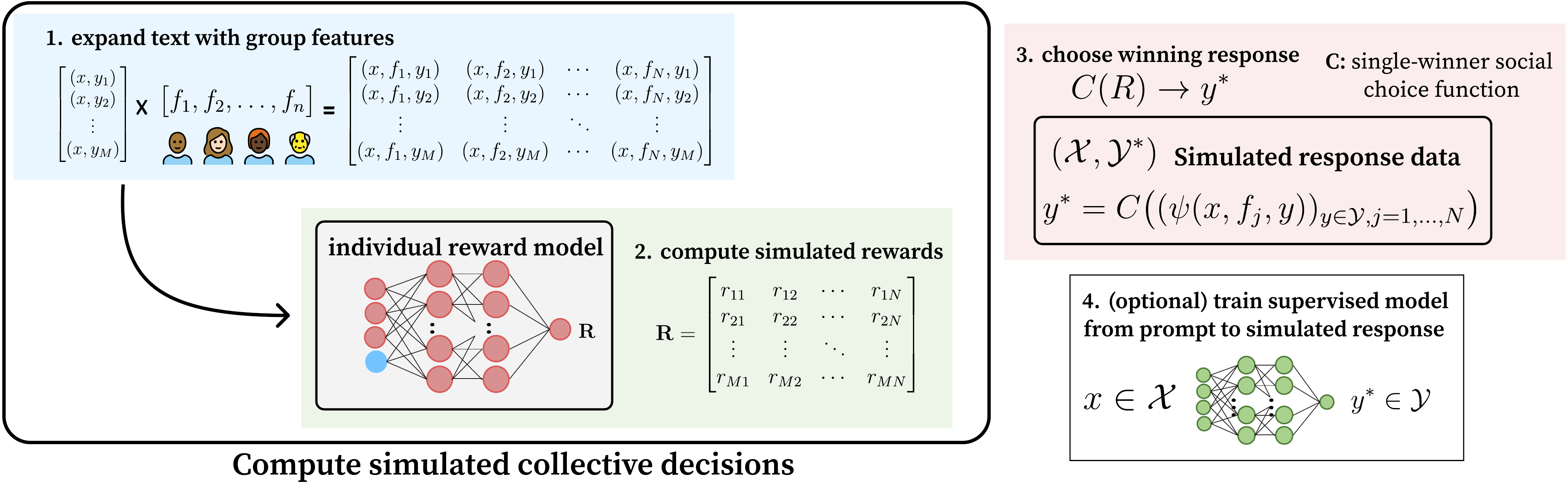}
\caption{\textbf{Supervised Learning from Simulated Collective Decisions}. We show that with an individual or cardinal reward model, as presented in \Cref{fig:rlchf-f}, responses $y$ to a prompt $x$ can be simulated.
This process expands the scope of studying preferences within RLHF and opens future work on personalization and other topics.}
\label{fig:slscd}
\end{figure*}

For one thing, one could modify ``RLCHF using evaluator features and aggregated ratings'' into ``Supervised Learning from Simulated Collective Decisions'', as shown in Fig.~\ref{fig:slscd}. For this, in step (3) from above, use the individual preference model $r_i = \psi(x,f_i,y)$ and feature vectors $f_1,\dots,f_N$ not to produce an aggregated rating but to simulate a collective choice that picks a single {\em winning} response $y^* = C\big( (\psi(x,f_j,y))_{y \in \mathcal{Y},j=1, \dots, N} \big)$. 
Here, $C$ is now a single-winner social {\em choice} function. Then in step (4), use data point $(x,y^*)$ to train the actual AI system via supervised (rather than reinforcement) learning. 
Instead of picking a single winner $y^*$, we could also use a multi-winner social choice function $C$ that outputs, say, a set of three responses $(y',y'',y''')$. These can then be (creatively) combined into a single response, e.g., 
%$y^* = L(y',y'',y''')$. 
by merging them into a bullet-point list and adding ``The following are (three) typical answers to your question: \dots'' at the beginning. 

A more radical modification would drop the fine-tuning-via-learning step altogether (leaving the LLM only pretrained) and instead simulate the collective choice at inference time. Whenever the live system is prompted with some $x$, generate $k\gg 1$ many candidate responses $y_i$ and $N\gg 1$ many evaluator feature vectors $f_j$ representative of the stakeholder population for the 
problem $(x,\mathcal{Y})$, and then directly return the winner $y^* = C\big( (\psi(x,f_j,y_i))_{j,i = 1}^{N,k} \big)$ 
of the simulated collective choice. Here, too, $C$ could be a multi-winner or probabilistic social choice rule.

\section{Which Traditional Social-Choice-Theoretic Concepts Are Most Relevant?}
\label{sec:rel SC concepts}

Social choice studies a wide variety of concepts, the relevance of which depends on the specific application. For example, consider the concept of {\em false-name-proofness}~\cite{Yokoo01:Robust,Yokoo04:Effect,Conitzer10:Using}, meaning that no-one can benefit from participating multiple times under multiple accounts. This is relevant when voting over the internet, but irrelevant for in-person faculty meetings where faculty vote publicly by raising their hands.

So, rather than studying every single social-choice-theoretic concept in the context of aligning AI systems, 
we should be careful to evaluate which traditional concepts are most relevant. In the following, we give just a few examples.

\subsection{Independence of Clones}
\label{sec:ind of clones}

In social choice problems, sometimes multiple alternatives, say $A$ and $B$, compare very similarly against every other alternative $X$, according to the preferences of individuals. Such alternatives are referred to as {\em clones}, a notion that can be formalized in several ways. According to a strict notion of clones \cite{Tideman87:Independence}, $A$ and $B$ are clones if, for every individual, if that individual prefers $A$ to some other alternative $X$, then they also prefer $B$ to $X$, and if they instead prefer $X$ to $A$, then they also prefer $X$ to $B$. According to a more liberal notion \cite{Laffond1996}, $A$ and $B$ are clones if, whenever a majority of individuals prefer $A$ to some other alternative $X$, then a majority prefers $B$ to $X$ as well, and whenever a majority prefers some $X$ to $A$, then a majority prefers $X$ to $B$ as well. 

Sometimes the introduction of a clone can affect the outcome of an election. Suppose a group of people are voting over where to go for dinner, and the only two alternatives are a Chinese restaurant and an Indian restaurant. 52\% of the voters prefer the Chinese restaurant. But then, someone points out that the Chinese restaurant has two floors and argues that the two floors should be considered separate options.
So now the alternatives are $C_1$, $C_2$, and $I$. Nobody really cares all that much about the floor, but suppose that 26\% of the voters prefer $C_1 \succ C_2 \succ I$, and 26\% of the voters prefer $C_2 \succ C_1 \succ I$ (adding up to the original 52\%). Further suppose that the voting rule used is Plurality, in which the alternative that appears at the very top of voters' rankings the most often wins.  Then the Indian restaurant ends up winning now with 48\% of the vote. This seems like an undesirable property for a voting rule to have; it would be better for the introduction of a clone never to make a difference.\footnote{More precisely, introducing a clone should not affect whether a non-clone (e.g., the Indian restaurant in our example) is selected or which non-clone is selected. But it may affect which clone, if any, is selected. For instance, a clone-independent rule could select $C_1$ over $C_2$ in our example, if among the 48\% of people who prefer $I$, a strict majority of them prefer $C_1$ to $C_2$.} This latter desirable property is called {\em independence of clones}. Perhaps when choosing restaurants, this is not that relevant, as restaurants will rarely be clones (unless the floors of restaurants are treated separately). On the other hand, when choosing responses for a chatbot, it may be quite common for two responses to be very close to each other, indicating its importance to this context. We note that Borda Count (see Figure \ref{Fig:diffagg}), which is implicitly used in some standard approaches to RLHF (\citealt{siththaranjan2023distributional}), badly violates independence of clones.

\subsection{Strategic Voting}
\label{sec:strat voting}

Another concern is {\em strategic voting} (or strategic feedback).
Strategic voting consists of casting a vote that does not reflect one's true preferences, in order to obtain a better result for oneself. For example, consider an election with plurality voting, as described above. A voter might perceive that her top-ranked alternative has no chance of winning and therefore strategically vote for another alternative.
Strategic voting poses a problem because we can no longer take votes (or feedback) at face value. Unfortunately, in general, every reasonable voting rule will sometimes introduce incentives to manipulate~\cite{Gibbard73:Manipulation,Satterthwaite75:Strategy}. These incentives to manipulate might be reduced if voters lack full information about the preferences of other voters \cite{Conitzer2011} or about the voting rule that will be used \cite{Holliday2019}. But we often cannot guarantee such ignorance, just as we often cannot guarantee computer security through obscurity.

What form might strategic voting in a context such as RLHF take? If rating responses range on a scale from (say) 0 to~10, a natural strategy is to overreport. E.g., if one evaluator does not really like a response (at the level of a 3), but suspects that others would like it (say, two other evaluators give it a 6), then this evaluator may strategically give a rating of 0 to ``compensate'' for the other reviewers. This manipulation would be successful if we eventually aggregate ratings by taking their average: the average will be pulled down to 4, instead of the 5 that would result from reporting truthfully, so that the average is closer to the 3 that the evaluator believes is ideal. If instead we use the median as the aggregate, then this manipulation is ineffective---the median would remain 6. Indeed, the median is {\em strategy-proof} in this context: misreporting one's preferences never helps, as long as one's only goal is to move the median rating closer to one's ``true'' rating (cf.~\citealt{Moulin1980}).

\subsection{Anonymity}

In democratic contexts, a standard desideratum on voting rules is \textit{anonymity}: if two voters swap their ballots before submitting them, the output of the voting rule will not change (the rules in Figure \ref{Fig:diffagg} all satisfy anonymity in this sense). This captures the idea that the voting rule should not favor some voters over others. Anonymity prohibits not only the extremes of dictatorship (recall Section~\ref{sec:background sc}) but even any kind of weighted voting wherein some voters' votes count for more than others. However, in the context of AI development, one might consider aggregating human feedback in a way that violates anonymity  (cf.~the {\em weighted majority rule} discussed in \citealt{NitzanParoush1982}). Perhaps some evaluators are more experienced or more highly rated than others; perhaps some are influenced by others, so their input should not be considered completely independent inputs for aggregation; and so on. In general, whether the same democratic norms applied to voting also apply in an AI context is an important question for discussion.

\subsection{Principles as Voters}
\label{sec:principles voting}

While it is standard in social choice for the voters to be humans, this is not required by the social choice theory framework. In some applications of social choice to AI ethics and safety, possibly including Constitutional AI (recall Section~\ref{sec:background const AI}), we might regard different ethical principles as the ``voters'' who can rank or otherwise evaluate the outputs of an AI system  (cf.~\citealt{Greene16:Embedding}). This is analogous to applications of social choice theory in the philosophy of science, where the ``voters'' are theoretical virtues that may rank scientific theories differently \cite{Okasha2011}, or to multi-criteria decision-making, where the ``voters'' are relevant factors that may rank the options differently \cite{Arrow1986}. Of course, such ethical principles could themselves be outputs of some prior social choice procedure in which the voters are humans (cf.~Collective Constitutional AI in \citealt{ganguli2023}).

This suggests a possible alternative architecture for applying social choice to AI---one sitting somewhere between the extremes of a spectrum that ranges from Constitutional AI at one end (in which principles are the whole show, while social choice does not appear) to the RLHF version of reinforcement learning as described in previous sections (where principles play no role at all). In this alternative model, each respondent would be required to justify her rankings of alternative AI responses in terms of their level of satisfaction of each of a number of principles taken from a fixed menu. The AI system would use the results to train for several independent tasks: for each principle, separately learn how to rate responses to queries based on that principle alone; and learn how to aggregate those separate ratings into an overall rating of the responses. These would be composed to form the final stage of a simulated collective decision---the stage in which the voters are the principles.

\section{How Should We Account for Behavioral Aspects and Human Cognitive Structures?}
\label{sec:behav aspects}

Preference elicitation often makes idealized assumptions, e.g., that each queried individual has well-defined and consistent preferences and will answer in a way that is  consistent with them (up to some random noise). But in reality, myriad behavioral effects kick in. For example, \citet{mcelfresh2021indecision} study how (in the context of kidney allocation) humans can become indecisive when asked to give input that will have moral implications.

This leads to a variety of questions about how to align AI systems in the context of behavioral effects. Should we correct for them? That would seem to require having a model where such a behavioral effect obscures humans' ``true" values. 
 But do these ``true'' values correspond to anything real in the world? Do we run the risk of the ``correction'' actually removing valuable information? Could the ability to make such ``corrections'' in fact be abused to intentionally remove inconvenient feedback?

\section{How Do We Navigate a Multiplicity of AIs?}
\label{sec:multi agent}

Consider, again, a group of people voting on a restaurant for their dinner. If there is a significant disagreement, then rather than forcing a minority to go someplace they really do not like, it can make sense to split into multiple groups, each going to their favorite restaurant. Similarly, perhaps it makes sense to create multiple AI systems---for example, to recognize strong inter- and intra-cultural variations that have been identified in some non-homogenous populations \cite{awad_moral_2018,PetersC2024}. 
Depending on the situation, the people providing feedback might be split into groups {\em ex ante} (e.g., each country makes a system based on their own citizens' feedback), but also {\em ex post}, where we first collect feedback and then form groups of people. The latter approach is closely related to the topic of {\em representation} in voting theory~\cite{faliszewski2017multiwinner}.

There is also the slightly different scenario where one AI system is in place, and some group of people believe that it is not serving them well. Hence, they might decide to pool their resources and create their own system. The literature on {\em cooperative game theory} 
(cf. \citealt{Chalkiadakis11:Computational}),  sometimes referred to as {\em coalitional game theory}, touches on these considerations (and indeed also plays a role in questions of representation, as in  %---see 
\citealt{Aziz17:Justified}).

Finally, let us highlight possible shortcomings to creating multiple AI systems. 
As in the restaurant example, it may have the result of unnecessarily dividing people into separate groups. Moreover, splitting into groups may not be feasible if it does not dovetail with existing social structures. For example, the US Federal Government may want to adopt a single system that will impact all its citizens, and adopting two systems would be tantamount to splitting the country in two.
Finally, unlike in the case of the restaurants, the multiple AI systems may have to interact with each other, creating the risk of conflict between AIs with different goals. The nascent literature on {\em cooperative AI}~\cite{Dafoe21:Cooperative,Conitzer23:Foundations} may help ensure these interactions do not result in adverse outcomes. Nonetheless, it might be best to see if we can completely avoid having multiple AIs with competing goals, or at least design them in a way that makes conflict between them less likely.

\section{Conclusion}
\label{sec:conclusion}

It is important that a variety of stakeholders are involved in giving input or feedback on how AI systems, such as those based on LLMs and other foundation models, should function. But those stakeholders are likely to give conflicting input. If so, how do we aggregate this input or otherwise use it for real or simulated collective decisions to end up with a sensible system? As we have argued in this paper, the field of social choice is well placed to help address this question---conceptually, due to its focus on methods for making consistent collective decisions, e.g., via aggregating preferences, judgments, and other inputs in a consistent way, as well as pragmatically, with many researchers in the computational social choice community being well prepared to engage with AI alignment researchers on these problems.

That said, it is important to acknowledge that aggregating conflicting input or feedback can be a complex task. It requires careful consideration of various factors, such as who the stakeholders are, which humans should provide the feedback, how their input is collected and weighed, the level of expertise and credibility of their input, and potential biases. Additionally, incorporating transparency and accountability measures into the aggregation process can help ensure that the final system reflects a fair and balanced representation of the stakeholders and their input. Significant research is needed to deepen our understanding of the possibilities of using social choice for these purposes and the different effects that this will have. 

Needless to say, the questions considered are multifaceted and cannot be adequately addressed without complementary (and not necessarily AI-specific) research. 
The involved practical decisions and associated legal and political considerations are important topics for future research as well.

Last but not least, we have put a particular focus on RLHF in this paper as it is an especially important and fruitful point of contact between social choice and AI. But the insights afforded by social choice theory bear on countless problems. Social choice can be used to more generally determine the objectives that AI systems pursue, the data on which they are trained, and which systems we build in the first place. Given the rapid development of AI systems underway, we urge researchers to begin forging these connections between social choice and AI alignment.

\section*{Impact Statement}

This paper highlights the need for further research and collaboration between experts in social choice and AI ethics and safety to ensure that AI systems are designed and deployed in a way that aligns with societal values and promotes accountability and transparency. 
As we discussed briefly in the introduction, we believe the approach proposed in this paper would result in systems that are fairer, that are less likely to have blind spots due to nobody having raised an issue, and that people buy into more broadly.  

As we briefly discussed in the introduction, there is perhaps a concern that feedback from a broader set of participants is more likely to be inconsistent and that consequently the resulting system will behave erratically.  The idea that na\"ive aggregation of votes or judgments leads to inconsistency is a familiar one from social choice theory.  For example, if three voters rank three alternatives $A$, $B$, and $C$ respectively as follows: $A \succ_1 B \succ_1 C$, $C \succ_2 A \succ_2 B$, and $B \succ_3 C \succ_3 A$, then a majority of voters prefers $A$ to $B$, a majority prefers $B$ to $C$, and a majority prefers $C$ to $A$.
This illustrates that majority preferences are cyclical and thus arguably irrational.   (See Figure~\ref{Fig:diffagg} for another example.)
We encounter similar issues in {\em judgment aggregation} (for an overview, see \citealt{EndrissChapter}).  To illustrate this in our own context, say that it is broadly agreed that an output should be given if and only if it is both safe and helpful.  Suppose evaluator 1 believes the output is safe but not helpful, and therefore should not be given.  Evaluator 2 believes the output is helpful but not safe, and therefore should not be given.  Evaluator 3 believes the output is both safe and helpful, and therefore should be given.  Then a majority believes that the output is safe, a majority believes that it is helpful, but a majority believes that it should not be given---so that majority judgments are logically inconsistent.  
However, social choice theory is precisely concerned with how we should actually obtain {\em consistent} aggregations and  is therefore well placed to address this issue.  For example, one common strategy is to restrict to rational or consistent outputs only and among these find one that is in some sense  ``closest'' to the reports (see, e.g., \citealt{ElkindSlinkoChapter}).  Therefore, social choice theory is actually well positioned to {\em help} with the issue of inconsistencies from aggregation.

\section*{Acknowledgements}

This paper grew out of the workshop on Social Choice for AI Ethics and Safety held at UC Berkeley in December 2023. We thank all the participants of the workshop for fruitful discussions. We are also grateful to Open Philanthropy for a grant that made the workshop possible, as well as further support from the Center for Human-Compatible AI, the C3.ai Digital Transformations Institute, and the Kavli Center for Ethics, Science, and the Public. For helpful comments on this paper, we thank Aditi Raghunathan and 
anonymous reviewers.


\bibliography{example_paper}
\bibliographystyle{icml2024}

%%%%%%%%%%%%%%%%%%%%%%%%%%%%%%%%%%%%%%%%%%%%%%%%%%%%%%%%%%%%%%%%%%%%%%%%%%%%%%%
%%%%%%%%%%%%%%%%%%%%%%%%%%%%%%%%%%%%%%%%%%%%%%%%%%%%%%%%%%%%%%%%%%%%%%%%%%%%%%%
% APPENDIX
%%%%%%%%%%%%%%%%%%%%%%%%%%%%%%%%%%%%%%%%%%%%%%%%%%%%%%%%%%%%%%%%%%%%%%%%%%%%%%%
%%%%%%%%%%%%%%%%%%%%%%%%%%%%%%%%%%%%%%%%%%%%%%%%%%%%%%%%%%%%%%%%%%%%%%%%%%%%%%%
% \newpage
% \appendix
% \onecolumn

% You can have as much text here as you want. The main body must be at most $8$ pages long.
% For the final version, one more page can be added.
% If you want, you can use an appendix like this one.  

% The $\mathtt{\backslash onecolumn}$ command above can be kept in place if you prefer a one-column appendix, or can be removed if you prefer a two-column appendix.  Apart from this possible change, the style (font size, spacing, margins, page numbering, etc.) should be kept the same as the main body.
%%%%%%%%%%%%%%%%%%%%%%%%%%%%%%%%%%%%%%%%%%%%%%%%%%%%%%%%%%%%%%%%%%%%%%%%%%%%%%%
%%%%%%%%%%%%%%%%%%%%%%%%%%%%%%%%%%%%%%%%%%%%%%%%%%%%%%%%%%%%%%%%%%%%%%%%%%%%%%%

\end{document}